\documentclass{jdmdh}
\usepackage{array}
\usepackage{pgfplots}
\pgfplotsset{compat=1.18}
\usepackage{multirow}

\titlespacing*{\section}
{0pt}{2.5ex plus 1ex minus .2ex}{1.3ex plus .2ex}
\titlespacing*{\subsection}
{0pt}{1.0ex plus 1ex minus .2ex}{1.3ex plus .2ex}
\titlespacing*{\subsubsection}
{0pt}{1.0ex plus 1ex minus .2ex}{1.3ex plus .2ex}

\title{Explaining Humour Style Classifications: An XAI Approach to Understanding Computational Humour Analysis}
\author[1]{Mary Ogbuka Kenneth}
\author[2]{Foaad Khosmood}
\author[1]{Abbas Edalat}
\affil[1]{Algorithmic Human Development group, Department of Computing, Imperial College London, UK } 
\affil[2]{Computer Engineering Department, California Polytechnic State University, USA} 

\corrauthor{Mary Ogbuka Kenneth}{m.kenneth22@imperial.ac.uk}

\begin{document}

\maketitle

\abstract{Humour styles can have either a negative or a positive impact on well-being. Given the importance of these styles to mental health, significant research has been conducted on their automatic identification. However, the automated machine learning models used for this purpose are black boxes, making their prediction decisions opaque. Clarity and transparency are vital in the field of mental health. This paper presents an explainable AI (XAI) framework for understanding humour style classification, building upon previous work in computational humour analysis. Using the best-performing single model (ALI+XGBoost) from prior research, we apply comprehensive XAI techniques to analyse how linguistic, emotional, and semantic features contribute to humour style classification decisions. Our analysis reveals distinct patterns in how different humour styles are characterised and misclassified, with particular emphasis on the challenges in distinguishing affiliative humour from other styles. Through detailed examination of feature importance, error patterns, and misclassification cases, we identify key factors influencing model decisions, including emotional ambiguity, context misinterpretation, and target identification. The framework demonstrates significant utility in understanding model behaviour, achieving interpretable insights into the complex interplay of features that define different humour styles. Our findings contribute to both the theoretical understanding of computational humour analysis and practical applications in mental health, content moderation, and digital humanities research.}

\keywords{Explainable AI; Computational Humour; Humour Style Classification; Natural Language Processing;  Machine Learning Interpretability; Digital Humanities}

\section{Introduction}


Humour is the tendency to experience or provoke laughter or provide amusement through written or spoken words \citep{Sen2012HumourResearch}. It plays a vital role in interpersonal interactions, emotional expression \citep{Amjad2022HumorAdults}, and psychological well-being \citep{Martin2003IndividualQuestionnaire, Chen2007AStudents,Edalat2023Self-initiatedLaugh,Martin2018TheApproach}. Different humour styles—self-enhancing, self-deprecating, affiliative, and aggressive—carry distinct emotional undertones and social implications \citep{Martin2003IndividualQuestionnaire,Kuiper2016IdentityWell-Being}. While affiliative humour fosters positive social interactions, aggressive humour may strain relationships through its potential to offend or demean \citep{Anderson2016AggressiveConflict}. Understanding these styles has significant implications across multiple domains, including mental health \citep{Edalat2024Self-InitiatedAgent, Martin2018TheApproach,Martin2003IndividualQuestionnaire}, content moderation \citep{Zhu2022AggressivePerspective, Sari2016WasHumor}, and artificial intelligence (AI) \citep{Kenneth2024SystematicClassification,KennethOgbuka2024ARecognition}. However, computational humour style recognition presents unique challenges in natural language processing (NLP) due to its subjective nature and complex psychological underpinnings \citep{Kenneth2024SystematicClassification,Kazienko2023Human-centeredHumor,Amjad2022HumorAdults}. Even though recent developments have been effective in categorising various humour styles, these models' decision-making procedures remain mainly unknown. Understanding how and why machine learning models categorise various humour styles is crucial for advancing both computational linguistics and digital humanities research \citep{Cortinas-Lorenzo2024TowardReview}.

Recent work by \citet{KennethOgbuka2024ARecognition} established baseline performance for humour style classification across four categories: self-enhancing, self-deprecating, affiliative, and aggressive humour. Their two-model approach achieved promising results, particularly with the General Text Embeddings Upgraded (ALI) + XGBoost for the single model configuration and the Multilingual E5 Text Embeddings (MUL) + XGBoost + ALI+XGBoost for the two-model configuration. However, the interpretability of these classifications—understanding which features and patterns drive the model's decisions—remains an open challenge.

Explainable AI (XAI) has emerged as a solution to address the opacity of traditional ML models by providing insights into their decision-making processes \citep{Lyu2024TowardsSurvey}. Through techniques such as Local Interpretable Model-Agnostic Explanations (LIME) \citep{Ribeiro2016WhyClassifier} and Shapley Additive Explanations (SHAP) \citep{Lundberg2017APredictions}, XAI enables researchers to identify key features driving predictions. This interpretability is especially crucial for recognising humour styles, as the interaction of linguistic patterns, emotional tones, and semantic nuances necessitates a thorough comprehension. For instance, identifying features such as sarcasm, sentiment contrasts, or emotions can provide more detailed information about why a certain text is categorised as affiliative or self-deprecating.

This paper extends the work of \citet{KennethOgbuka2024ARecognition} by introducing an XAI framework for humour style classification. By applying XAI to the best-performing single model (ALI+XGBoost) reported in their research, we provide detailed interpretability that highlights the influence of linguistic and emotional features on model predictions. This approach not only makes the classification of humour styles more transparent, but it also gives researchers practical insights that allow them to further investigate the role of humour in communication and psychological well-being.

The primary contributions of this paper are as follows:
    \begin{enumerate}
        \item Development of a comprehensive XAI framework tailored to humour style recognition.
        \item Detailed analyses of how linguistic and affective features contribute to humour style classification.
        \item Insights into the practical implications of these findings for researchers studying humour and its applications.
    \end{enumerate}

By addressing the critical need for explainability, this study bridges the gap between computational performance and interpretability in the field of humour style recognition.

\section{Related Works}

The development of explainable approaches to computational humour analysis converges three research streams: general XAI methodologies for text classification, interpretable humour analysis models, and explainable style classification approaches. This section examines these areas' contributions to computational humour recognition interpretability, progressing from foundational XAI methods to specific style-based classifications.

\subsection{General XAI Methods for Text Classification}
Recent advances in XAI for text classification demonstrate varied approaches to model interpretability. \citet{Perez-Landa2021AnTweets} combined emotional, sentiment, syntactic, and lexical features for xenophobic content detection, achieving F1-scores of 76.8\% and 73.4\% on different datasets. Although their keyword-based pattern matching may face generalisation challenges, their success with emotion and sentiment features informs our approach to humour style explanation.

\citet{Chowdhury2021ExplainingLIME} applied LIME to explain black-box sentiment analysis models, achieving 72\% accuracy with Long Short-Term Memory (LSTM) networks and FastText embeddings. Their implementation revealed word-level contributions to sentiment predictions through probability scores and contextual relationships, demonstrating LIME's effectiveness for deep learning interpretability.

\citet{Ahmed2022ExplainableDisorder} integrated attention mechanisms with fuzzy logic rules for interpretable sentiment analysis, achieving 89\% F1-score. Their dual-layer approach combined local explanations with global interpretability through attention weights and fuzzy rules. While effective for mental health applications, questions remain about adapting such hybrid systems for subjective tasks like humour classification, where decision boundaries may be less clearly defined.

In humour-adjacent tasks, \citet{Ortega-Bueno2022Multi-viewVariants} employed attention mechanisms for irony detection in Spanish, demonstrating that different attention types focus on distinct linguistic features. However, their approach prioritised performance over explainability, underscoring the need for XAI methods that consider both linguistic feature utilisation and cross-cultural variations.

\subsection{Interpretable Models for Humour Analysis}
Early interpretable humour analysis focused on transparent feature development. \citet{Zhang2017InvestigationsRecognition} proposed interpretable features for humour recognition: contextual knowledge (modelling semantic relationships), affective polarity (quantifying emotional impact), and subjectivity (capturing personal interpretation). Their CASHCF (Combined All Semantic and Human-Centred Features) model improved performance while maintaining transparency but highlighted challenges in capturing deep semantic relationships in an explainable manner.

\citet{Mann2024CLEFClassification} addressed pun detection, humour classification (irony, exaggeration, incongruity, self-deprecating, wit), and English-French joke translation using TF-IDF features and MarianMT. They found wit the easiest to classify, while irony and exaggeration were more challenging, supporting our style-specific XAI approach. However, their reliance on basic features limits interpretability. Our framework advances this by incorporating semantic, emotional, and linguistic analyses.

\citet{DeMarez2024THInC:Detection} introduced THInC (Theory-driven Humour Interpretation and Classification), a framework grounding humour detection in psychological theories. Using GA2M classifiers aligned with humour theories (superiority, relief, incongruity, and incongruity resolution), they achieved an F1 score of 85\% through theory-informed proxy features like emotional bursts and sentiment shifts. Their mapping of computational features to theoretical constructs informs our XAI approach, though their focus is on binary humour classification rather than style differentiation.

\citet{Mahajan2024AnModels} proposed a humour detection model using ensemble learning and Berger's humour typology features (e.g., emotive, incongruity, intensity). Their stacking-based ensemble achieved 85.68\% accuracy and a 72.57\% F1-score on Yelp reviews, with ablation studies highlighting the value of theory-driven features. While focused on binary detection, their use of typology offers insights for interpretable humour analysis.

\citet{Chen2024TalkInterpretation} introduced TalkFunny, a Chinese humour response dataset with "chain-of-humour" annotations and mind maps explaining response generation. Their PLM-based framework, integrating interpretability tools, showed improved performance. While focused on humour generation, their approach supports the value of explainable components, aligning with our goals for interpretability in humour AI.

\subsection{Explainable Approaches to Style Classification}

Style classification has evolved from binary to multi-class approaches. 
\citet{Abulaish2018Self-DeprecatingApproach} developed a two-layer system for self-deprecating sarcasm detection (94\% F1-score), using explicit linguistic rules and feature-based classification. Despite its success in binary classification, their work underscored the need for more advanced explainability to handle multiple humour styles. \citet{Kamal2020Self-deprecatingApproach} expanded this work with three explicit feature categories (self-deprecating patterns, exaggeration markers, and word embeddings) for self-deprecating humour detection, achieving F1-scores of 62\%–87\% across datasets. 

\citet{KennethOgbuka2024ARecognition} introduced a dataset and a two-model approach for humour style classification, targeting four styles (self-deprecating, self-enhancing, affiliative, aggressive) and non-humour. They addressed challenges in distinguishing affiliative and aggressive humour by using a four-class classification followed by binary discrimination, achieving a 78.6\% F1-score with improved differentiation. However, their work left unanswered questions about misclassification reasons between styles—a gap our XAI framework seeks to fill. 

While significant progress has been made in humour classification and XAI techniques, two significant gaps persist: most interpretable approaches focus on binary rather than style-specific classification, and current style-specific approaches lack comprehensive explanatory frameworks for linguistic and emotional feature interaction. Our framework addresses these gaps through a unified approach to humour style classification explanation, supporting both theoretical understanding and practical applications. A comprehensive summary of related works and their contributions is
presented in 
Table \ref{Table:related_works} (Appendix \ref{appendix:related_works}).

\section{Methodology}
This study develops an XAI framework to analyse humour style classification by examining model predictions, linguistic features, and emotional patterns. Our methodology comprises three main components: dataset and model selection, prediction analysis using LIME, and comprehensive feature analysis across linguistic, affective, and contrast patterns.

\subsection{Dataset and Classification Model}
This study utilises the dataset and ALI+XGBoost model from \citet{KennethOgbuka2024ARecognition}, which achieved 77.8\% accuracy and 77.3\% F1-score. The dataset comprises 1,463 instances gathered from multiple sources:
\begin{itemize}
    \item 983 jokes from different websites (Reader's Digest, Parade, Bored Panda, Laugh Factory, Pun Me, Independent, Cracked, Reddit, Tastefully Offensive, and BuzzFeed), labelled based on original website tags and humour theory.
    \item 280 non-humorous text instances from the ColBERT dataset \citep{Annamoradnejad2020ColBERT:Humor}
    \item 200 instances from the Short Text Corpus \footnote{Short Text Corpus (\url{https://github.com/CrowdTruth/Short-Text-Corpus-For-Humor-Detection})}, including 150 jokes and 50 non-jokes
\end{itemize}

To ensure annotation quality and mitigate potential biases, six Ph.D. candidates from Africa, Asia, and Europe independently annotated 200 randomly selected instances from the Short Text Corpus. Each set of 100 samples was reviewed by three annotators, with final labels determined by majority vote.
The dataset contains five categories: self-enhancing (298 instances), self-deprecating (265), affiliative (250), aggressive (318), and neutral humour (332). Text lengths vary considerably, ranging from 4 to 229 words, with a total vocabulary of 4,506 unique words. Each humour style has the following linguistic characteristics:

\begin{itemize}
    \item \textbf{Self-enhancing:} 1,181 unique words, with lengths ranging from 5 to 35 words, median of 13 and mode of 10 words
    \item \textbf{Self-deprecating:} 1,365 unique words, with lengths ranging from 4 to 148 words, median of 15 and mode of 12 words
    \item \textbf{Affiliative:} 1,376 unique words, with lengths ranging from 5 to 229 words, median of 15 and mode of 13 words
    \item \textbf{Aggressive:} 1,449 unique words, with lengths ranging from 4 to 112 words, median of 15 and mode of 12 words
    \item \textbf{Neutral:} 1,856 unique words, with lengths ranging from 4 to 34 words, median of 10 and mode of 9 words
\end{itemize}

While the length distributions show positive skewness with some long examples (particularly in affiliative humour), the central tendencies are quite similar across humour styles, with medians ranging from 10-15 words and modes from 9-13 words. The neutral category shows the most compact distribution (4-34 words), while humorous categories occasionally include longer examples. However, since the majority of instances cluster around similar lengths, these outliers are unlikely to impact classification performance significantly.

To illustrate these categories, representative examples from the humour styles dataset are presented in Figure \ref{fig:examples_styles}:

\begin{figure}[h]
    \caption{Examples for Each Category in the Dataset}
    \label{fig:examples_styles}
    \includegraphics[width=\columnwidth]{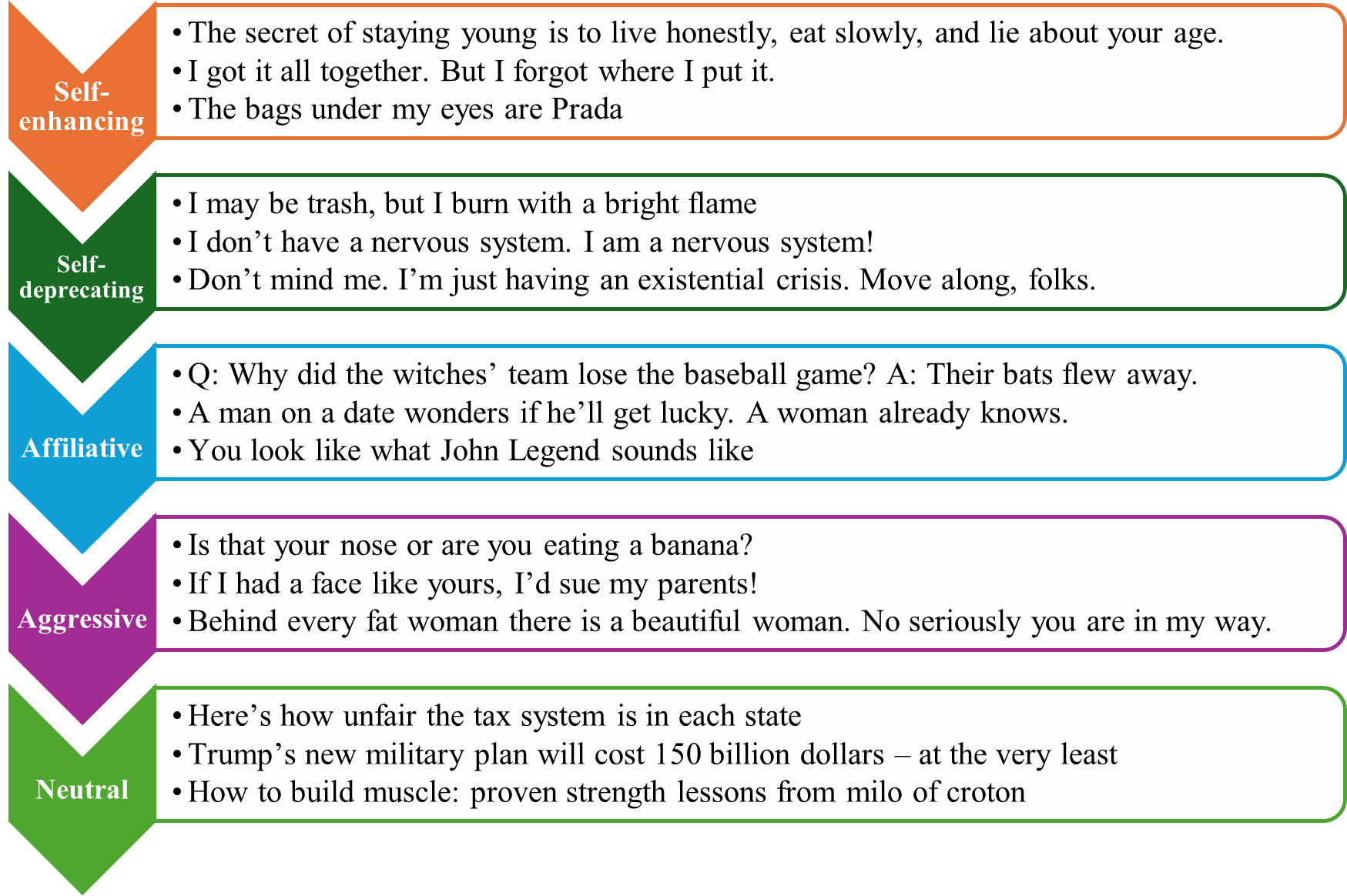}
\end{figure}

Using this dataset, \citet{KennethOgbuka2024ARecognition} developed two classification approaches: a single-model and a two-model system. The single-model directly classified input text into one of the five categories (self-enhancing, self-deprecating, affiliative, aggressive, and neutral). Due to challenges in distinguishing affiliative from aggressive humour, they developed a two-model approach that:
\begin{enumerate}
    \item First classifies text into four categories (self-enhancing, self-deprecating, neutral, and a combined affiliative/aggressive class).
    \item Then uses a separate binary classifier to distinguish between affiliative and aggressive instances for texts classified in the combined class.
\end{enumerate}

In this study, we analyse the prediction of the best-performing single-model (ALI+XGBoost) to understand these classification challenges from the perspective of explainable AI.

\subsection{Model Prediction Analysis}
We employ LIME (Local Interpretable Model-agnostic Explanations) due to its advantages for word embedding-based classification \citep{Ribeiro2016WhyClassifier}. LIME provides word-level interpretability through local explanations while maintaining model-agnostic analysis and computational efficiency. Our analysis progresses through:
\begin{enumerate}
    \item Individual prediction explanation using LIME
    \item Feature importance extraction from XGBoost
    \item LIME result visualization
    \item Analysis of confidence score across humour styles
\end{enumerate}

\subsection{Feature Analysis}
We categorise the patterns extracted to explain the classification model into three groups: linguistic patterns, affective patterns, and contrast patterns. Our analysis leverages multiple feature detection algorithms, each with different capabilities and limitations.

\subsubsection{Linguistic Patterns}
Linguistic patterns capture the language elements that contribute to humour's comedic effect. These patterns manipulate structure, context, and style to create humorous situations \citep{Kenneth2024SystematicClassification}. The key linguistic patterns analysed include:

\paragraph{Sound Patterns } Sound patterns exploit auditory features of language, such as rhyme, alliteration, and homophony.
\begin{itemize}
    \item \textbf{Rhyme}: The repetition of similar sounds at the ends of words. Examples include \textit{sight} and \textit{flight}, \textit{sad} and \textit{mad}, \textit{cat} and \textit{hat}. In our analysis, we used the ``pronouncing" library with the CMU Pronouncing Dictionary to detect rhymes. Although the dictionary effectively covers standard English pronunciations, it may not capture slang or neologisms present in the dataset. For instance, out of the 4,506 unique words in the dataset, 90.75\% (4,089 words) were found in the CMU Pronouncing Dictionary. Unmatched examples include: \textit{`launchalot', `aronofskys', `500000', `ahappy', `admited', `kanye', `houseplant', `idk', `behaviours'}
    
    \item \textbf{Alliteration}: The repetition of initial consonant sounds in closely placed words, such as in \textit{"Crazy cats create chaos"}. Implemented using NLTK’s phonetic detection and spaCy’s tokenisation, the algorithm identifies words sharing initial phonemes via the CMU Pronouncing Dictionary. However, this approach has limitations:
    \begin{itemize}
        \item It considers only the first phoneme, occasionally grouping words incorrectly (e.g., \textit{"when"} and \textit{"one"}).
        \item It counts repeated words as alliteration.
        \item It ignores word proximity, leading to false positives.
    \end{itemize}
    \item \textbf{Homophones}: Words that sound identical but have different meanings or spellings. Examples include: \textit{cell} and \textit{sell}, \textit{sea} and \textit{see}, \textit{to}, \textit{too}, and \textit{two}. Detection was implemented using \href{https://wordhoard.readthedocs.io/en/latest/basic_usage/}{WordHoard’s} homophone identification combined with WordNet synsets for semantic differentiation. While this approach performed well, it remains limited by WordNet's vocabulary size.
\end{itemize}

\paragraph{Wordplay} This category analyses deliberate manipulations of word meanings to create humour.
\begin{itemize}
    \item \textbf{Puns}: Words with similar pronunciations but different meanings, identified through phonetic similarity and semantic difference analysis. For instance, \textit{"A bicycle can't stand on its own because it is \textbf{two-tired}."} Here, the bolded word is a homophonic pun, replacing too-tired with two-tired. Our implementation detects homophonic puns using phonetic matching (using `pronouncing') and meaning differentiation (using WordNet). Current limitations include:
    \begin{itemize}
        \item Only detects single-word homophonic puns with exact sound matches
        \item Cannot identify near-homophones, multi-word puns, or contextual wordplay
        \item Misses puns based on multiple meanings of the same word
    \end{itemize}
    \item \textbf{Synsets}: Words or phrases that can be interpreted in multiple ways, detected using WordNet synset analysis. Each word in the dataset is matched with its WordNet synsets, extracting all possible meanings. Approximately 88.50\% (3,988 of 4,506 words) of the vocabulary had synset coverage. Words like \textit{`ourselves', `launchalot', `aronofskys', `since',} and \textit{`500000'} were not included in WordNet.
    \item \textbf{Syllabic Structure}: Patterns in syllable counts and distributions within the text, analysed for their role in creating rhythm or emphasis. This was implemented using the Pyphen dictionary's hyphenation patterns as a proxy for syllable counting. In this approach, the algorithm counts hyphenation points in each word using Pyphen's default hyphenation rules. This method has inherent limitations, as hyphenation patterns do not always correspond directly to syllabic boundaries. Potential limitations include:
    \begin{itemize}
        \item Reliance on hyphenation rules rather than true syllabic structure
        \item No special handling for compound words or complex morphology
        \item Limited accuracy for abbreviations, numerals, and non-standard English words
        \item Dependency on Pyphen's dictionary coverage
    \end{itemize}
\end{itemize}

\paragraph{Structural Elements} 
Structural elements focus on how linguistic arrangements contribute to humour.
\begin{itemize}
    \item \textbf{Self-references}: Details with the identification of first-person pronouns and personal narratives. We used basic tokenisation to identify explicit self-references (\textit{`i', `me', `my', `mine', `myself'}) and associated context. However, implicit self-references may be missed.
    \item \textbf{Parts of speech (POS)}: Analysis of the distribution and arrangement of word types, such as nouns, verbs, and adjectives. POS tagging was implemented using TextBlob, which offers basic functionality but lacks handling for informal language or social media text. 
    \item \textbf{Grammar Complexity}: Measurement of sentence complexity by analysing the frequency of specific syntactic dependencies, such as clausal complements (ccomp), open clausal complements (xcomp), and adverbial clause modifiers (advcl).
\end{itemize}

These linguistic patterns were systematically extracted using NLTK, WordNet, \href{https://wordhoard.readthedocs.io/en/latest/basic_usage/}{wordhoard}, and custom phonetic analysis algorithms. By detecting these patterns, the analysis explains how different linguistic devices contribute to humour styles, providing interpretable features for the classification model.

\subsubsection{Affective Patterns}
Affective patterns relate to feelings, moods, and attitudes, as defined by the \textit{Collins English Dictionary}. In humour style analysis, these patterns help distinguish whether a statement is humorous, sarcastic, or offensive \citep{Kenneth2024SystematicClassification}. Key affective patterns analysed include:

\paragraph{Sarcasm} Sarcasm detection was performed using a \href{https://huggingface.co/jkhan447/sarcasm-detection-RoBerta-base-POS}{RoBERTa-based} model trained for sarcasm identification, achieving 60.7\% accuracy. The model provides binary classification (sarcastic/non-sarcastic) and probability scores, capturing subtle verbal irony often found in specific humour styles.

\paragraph{Sentiment} Sentiment analysis employed a multi-layered approach using \href{https://huggingface.co/siebert/sentiment-roberta-large-english}{RoBERTa} (93.2\% average accuracy) \citep{Hartmann2023MoreAnalysis} and TextBlob models to assess: 

\begin{itemize}
\item Dominant sentiment (positive, negative, neutral) with confidence scores
\item Sentiment strength measured through positive-negative score differential
\item Polarity (-1 to 1) indicating sentiment direction and intensity
\item Subjectivity and objectivity metrics to gauge emotional content
\end{itemize}

\paragraph{Emotion} Emotional content was analysed using a  \href{https://huggingface.co/bhadresh-savani/distilbert-base-uncased-emotion}{DistilBERT-based} model that classifies text into six discrete emotion categories (joy, anger, sadness, fear, love, and surprise) with an accuracy of 93.8\% and F1-score of 93.79\%. This fine-grained emotional analysis reveals distinctive affective patterns across different humour styles.

\subsubsection{Contrast Patterns}
Our analysis of contrast patterns focuses on semantic and sentiment-based contradictions that generate humorous effects through the following components:
\paragraph{Sentiment Contrasts} We examined opposing emotional valences at two distinct levels:
\begin{itemize}
\item \textbf{Sentence-level contrasts}: Sequential sentences were analysed for opposing sentiment polarities, capturing dramatic shifts in emotional tone.
\item \textbf{Word-level contrasts}: Content words (adjectives, adverbs, verbs, and nouns) were evaluated using SentiWordNet to identify internal sentiment conflicts within sentences.
\end{itemize}

\paragraph{Semantic Elements} We investigated semantic relationships and conflicts through:
\begin{itemize}
\item \textbf{Exaggeration markers}: Systematic identification of absolute terms and extreme descriptors that amplify narrative elements.
\item \textbf{Intensification patterns}: Analysis of linguistic intensifiers that heighten semantic impact.
\item \textbf{Semantic incongruity}: Quantification of conceptual contradictions using WordNet similarity metrics to identify semantically distant word pairs.
\end{itemize}

 Having established our methodological framework and feature detection capabilities, we now turn to the analysis of our results and their implications for humour style classification.

\section{Results and Discussion}
This section presents an analysis of the ALI+XGBoost model's performance in humour style classification, followed by a detailed examination of linguistic mechanisms and error patterns. We first evaluate the model's classification metrics and confidence scores across different humour styles, then analyse the distinctive linguistic features characterising each style, and finally examine misclassification patterns to understand the model's limitations. The complete LIME visualisations, linguistic analysis, and
affective analysis results for all 293 examined instances are available in our public
repository \footnote{XAI humour styles : \url{https://github.com/MaryKenneth/XAI_humour_styles/tree/main}}. 

\subsection{Model Performance Analysis}
The performance analysis of the ALI+XGBoost model highlights distinct patterns in classification accuracy and confidence across different humour styles. The model achieved an overall accuracy of 78\% and a macro-average F1-score of 77\%, with style-specific variations:
\begin{itemize}
    \item Self-enhancing humour exhibited the highest precision (0.82) and F1-score (0.83).
    \item Neutral content attained the highest recall (0.93) and a strong F1-score (0.85).
    \item Affiliative humour had the lowest recall (0.58), underscoring challenges in identifying this style.
    \item Aggressive humour showed balanced precision (0.75) and recall (0.74).
\end{itemize}

Confidence scores also varied significantly across styles (p \textless 0.0001): Self-enhancing humour had the highest average confidence (0.889), followed by neutral (0.887), self-deprecating (0.811), aggressive (0.780), and affiliative humour (0.748).

\subsection{ Linguistic Mechanism Analysis}
Analysis of linguistic mechanisms revealed distinct patterns across humour styles, as summarised in Table \ref{Table:complexity_statistics}. We examined four key dimensions: syllabic complexity, semantic conflicts, homonym usage, and exaggeration patterns.

\subsubsection{Syllable Complexity} 
Neutral content exhibited significantly higher syllabic complexity (mean: 1.355, SD: 0.245), suggesting more formal or complex language use. In contrast, aggressive humour demonstrated the lowest complexity (mean: 1.148, SD: 0.118), indicating a preference for simpler, more direct language. Other humour styles showed intermediate complexity levels, with self-enhancing (mean: 1.206) and self-deprecating (mean: 1.192) humour showing similar patterns.

\subsubsection{Semantic and Structural Elements}
Affiliative humour exhibited the highest frequency of semantic conflicts (mean: 29.755, SD: 114.755), significantly surpassing other styles. This suggests frequent use of wordplay and unexpected combinations. Self-deprecating humour followed with a mean of 18.717 (SD: 47.830). In contrast, neutral and self-enhancing content demonstrated markedly lower frequencies of semantic conflicts (mean: 7.025, SD: 7.347, and mean: 7.557, SD: 7.274, respectively).

Homonym usage patterns revealed significant differences between humorous and neutral content. Social forms of humour—affiliative and aggressive—showed the highest homonym frequencies (mean = 6.796, SD = 4.743 and mean = 6.667, SD = 3.772, respectively), while neutral content demonstrated significantly lower usage (mean = 2.662, SD = 1.987). This indicates that wordplay involving multiple word meanings is a common feature across humour styles.

\begin{table}[ht!]
\centering
\resizebox{1.0\textwidth}{!}{
    \begin{tabular}{l|llll|llll|llll|llll}
    \hline
    \textbf{} & \multicolumn{4}{l|}{\textbf{Syllable Complexity}}           & \multicolumn{4}{l|}{\textbf{Semantic Conflict Count}}       & \multicolumn{4}{l|}{\textbf{Homonym Count}}                & \multicolumn{4}{l}{\textbf{Exaggeration Count}}           \\ \hline
    \textbf{Predicted Class} & \textbf{Mean} & \textbf{Std} & \textbf{Min} & \textbf{Max} & \textbf{Mean} & \textbf{Std} & \textbf{Min} & \textbf{Max} & \textbf{Mean} & \textbf{Std} & \textbf{Min} & \textbf{Max} & \textbf{Mean} & \textbf{Std} & \textbf{Min} & \textbf{Max} \\ \hline
    Affiliative              & 1.183         & 0.118        & 1.0          & 1.48         & 29.755        & 114.755      & 0            & 794          & 6.796         & 4.743        & 1            & 33           & 1.388         & 1.304        & 0            & 5            \\ \hline
    Aggressive               & 1.148         & 0.118        & 1.0          & 1.50         & 13.439        & 37.424       & 0            & 231          & 6.667         & 3.772        & 2            & 22           & 1.456         & 1.593        & 0            & 9            \\ \hline
    Neutral                  & 1.355         & 0.245        & 1.0          & 2.00         & 7.025         & 7.347        & 0            & 48           & 2.662         & 1.987        & 0            & 11           & 0.738         & 0.853        & 0            & 3            \\ \hline
    Self-deprecating         & 1.192         & 0.171        & 1.0          & 2.00         & 18.717        & 47.830       & 0            & 294          & 6.087         & 3.817        & 0            & 22           & 1.848         & 1.738        & 0            & 8            \\ \hline
    Self-enhancing           & 1.206         & 0.143        & 1.0          & 1.64         & 7.557         & 7.274        & 0            & 30           & 4.902         & 2.350        & 0            & 12           & 0.885         & 0.950        & 0            & 4            \\ \hline
    \end{tabular}
    }
\caption{Complexity Statistics}
\label{Table:complexity_statistics}
\end{table}

\subsubsection{Rhetorical Devices}
Exaggeration patterns varied significantly across styles. Self-deprecating humour exhibited the highest frequency of exaggeration (mean = 1.848, SD = 1.738), followed by aggressive humour (mean = 1.456, SD = 1.593). Neutral and self-enhancing content showed significantly lower exaggeration frequencies (mean = 0.738, SD = 0.853 and mean = 0.885, SD = 0.950, respectively).

These findings suggest that humour styles employ distinctive combinations of linguistic mechanisms. Social forms of humour (affiliative and aggressive) predominantly utilise wordplay devices, while self-directed styles demonstrate varied complexity patterns. Neutral content maintains higher linguistic complexity but shows reduced usage of rhetorical devices.

\subsubsection{Mechanism Correlations and Interactions}
\begin{figure}[h]
    \includegraphics[width=\columnwidth]{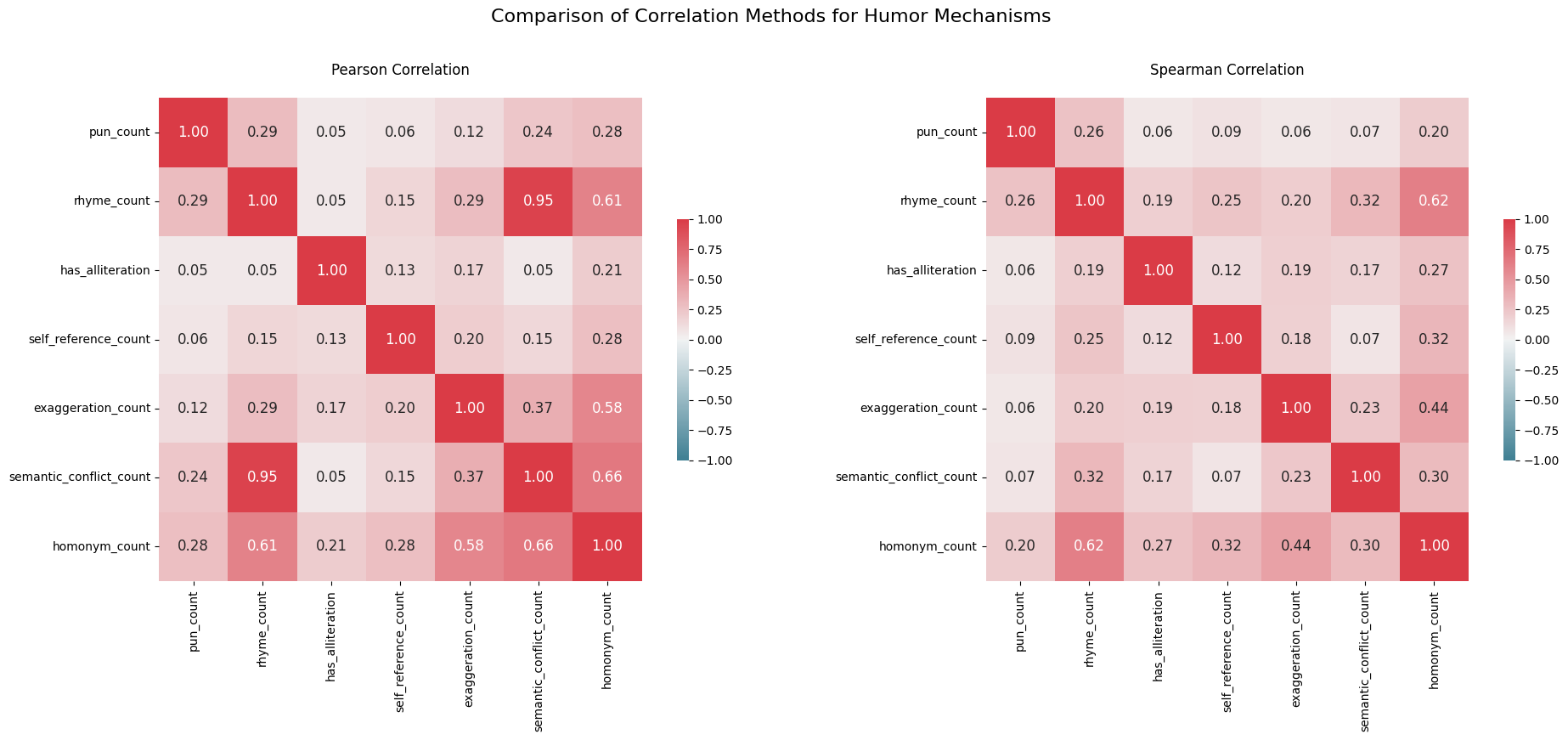}
    \caption{Mechanism Pearson and Spearman Correlations}
    \label{fig:pearson_spearman_correlation}
\end{figure}

Analysis of correlations between linguistic mechanisms revealed complex interaction patterns in humour construction. We employed both Pearson and Spearman correlation analyses as shown in Figure \ref{fig:pearson_spearman_correlation} to understand these relationships more thoroughly, as Spearman correlations are more robust to outliers and can detect non-linear relationships. 

\textbf{Correlation Disparity and Outlier Effects} \\ 
The most striking finding emerged in the disparity between Pearson and Spearman correlations, particularly for semantic conflict relationships. While Pearson correlation suggested a strong linear relationship between semantic conflicts and rhyme (r = 0.95), the much lower Spearman correlation (r = 0.32) revealed this relationship to be heavily influenced by outliers. This disparity manifests across different scaling levels:
\begin{itemize}
    \item \textbf{High-density outliers:} 794 conflicts/181 rhymes; 294 conflicts/44 rhymes; 231 conflicts/27 rhymes
    \item \textbf{Mid-range instances:} 48 conflicts/7 rhymes; 30 conflicts/5 rhymes
    \item \textbf{Typical cases:} Majority fall below 30 conflicts; Variable rhyme counts (0-5)
\end{itemize}
This distribution pattern is illustrated by two contrasting joke examples:

\textbf{A typical case example: }\textit{``Reminder that Winnie the Pooh wore a crop top with no pants and ate his fave food and loved himself. So you can too."} 
\begin{itemize}
    \item 4 rhyming pairs: \textit{(``Pooh/you", ``Pooh/too", ``crop/top", ``you/too")}
    \item 5 semantic conflict: memory aid vs physical action \textit{(``reminder/wore")},  memory aid vs emotional satisfaction \textit{(``reminder/loved")}, clothing vs eating \textit{(``wore/food"),} physical vs emotional satisfaction \textit{(``food/loved")}, and the phrase \textit{``So you can too"} suggests an invitation, but applying it to ``food" and ``loved himself" creates ambiguity \textit{(``food/too")}.
\end{itemize}

 \textbf{High rhyme example:}
\textit{``This is your mother. I just texted you but I don't know how to make the facey-things so... happy face at the end."}
\begin{itemize}
    \item 4 distinct rhyming pairs \textit{(``you/do", ``you/to", ``do/to", ``know/so")}
    \item No significant semantic conflicts
\end{itemize}

\textbf{Stable Correlations and Robust Relationships}\\
While many correlations weakened under Spearman analysis, some relationships demonstrated stability across both methods. Most importantly, the relationship between rhyme and homonym remained consistent (Pearson: 0.61, Spearman: 0.62), suggesting a robust underlying connection independent of outliers. Homonyms also showed stable correlations with exaggeration (Pearson: 0.58, Spearman: 0.44) and semantic conflicts (Pearson: 0.66, Spearman: 0.30).

Consider this example that demonstrates stable mechanism interactions:
\textit{``I forgot you can't make depression jokes outside of twitter Imao my coworker was like 'you ready for this year to be over?' I was like I'm ready for this life to be over' he was like bro what"}.
This joke exhibits stable correlations between multiple features:
\begin{itemize}
    \item 7 rhyming pairs such as \textit{``I/my", ``you/to", ``be/he"}
    \item 6 homonyms like \textit{{``I": [``eye", ``aye", ``ai"]; `you": [``yew", ``ewe"]}}
    \item 10 semantic conflicts supporting wordplay such as `\textit{`forgot/depression", ``forgot/year", ``forgot/life", ``make/year"}
    \item 3 self-references integrated with sound patterns-\textit{``I/my/I"}
    \item 12 alliteration instances like \textit{{``F": [``forgot", ``for"]; ``Y": [``you", ``year"]; ``M": [``make", ``my"]; ``T": [``Twitter", "to"]}}
\end{itemize} 
\textbf{} \hfill \break
\textbf{Independent Mechanisms}\\
Several mechanisms demonstrated operational independence. Self-referential content maintained weak to moderate correlations with other features --Pearson range:\begin{math}(0.06 \le r \le 0.28)\end{math} and similar stability range for Spearman, suggesting it functions largely autonomously in humour construction. Similarly, alliteration showed minimal correlation with other mechanisms (Pearson correlations \textless 0.21 and Spearman correlations \textless  0.27 across all comparisons), indicating its role as an independent supplementary device rather than a core component of humour construction.\\
Example joke: \textit{``My manager asked if I take constructive criticism and I said yes while wiping away my teary eyes."}

In the above joke, self-reference is present in four instances: \textit{``my/manager", ``I/take", ``I/said", and ``my/teary"}. These create humour through a personal narrative, establishing a relatable and emotional context without requiring support from other mechanisms.

Alliteration appears in six groups, such as the \textit{``M"} sounds in \textit{[``my," ``manager," ``my"] }and the \textit{``T"} sounds in \textit{[``take," ``teary"]}. While alliteration enhances the rhythmic flow of the joke, it functions as a supplementary device, adding aesthetic value without depending on other features for its effect.

The semantic structure of the joke contains 25 semantic conflicts such as \textit{``manager/criticism"}, \textit{``manager/teary"}, and \textit{``manager/yes"}, which operate independently of sound patterns. Looking at these three example conflicts: 1) the manager's formal criticism meeting an emotional response - demonstrating the professional-personal divide that operates independently of phonetic features; 2) the stark contrast between professional authority and teary vulnerability - showcasing how semantic conflicts can generate humour without relying on sound patterns; and 3) the ironic juxtaposition of verbal affirmation against visible distress - illustrating how meaning-based conflicts drive the humour independently of linguistic devices like rhyme or wordplay.

This example illustrates how different humour mechanisms (self-reference, alliteration, and semantic conflict) can function independently, each contributing uniquely to the joke's overall impact.\\
Pun usage demonstrated selective associations, showing moderate correlations with sound-based features including rhyme (Pearson: 0.29, Spearman: 0.26) and homonyms (Pearson: 0.28, Spearman: 0.20). However, correlations with other mechanisms were weaker (Pearson(r) \textless  0.25, Spearman (r) \textless  0.10). This pattern suggests that while puns integrate sound patterns and multiple meanings, they operate through relatively distinct linguistic pathways in humour generation.

\textbf{Mechanism Correlation Conclusion}\\
The comparison between Pearson and Spearman correlations reveals that humour mechanisms interact through multiple parallel pathways rather than a single unified system. While some mechanisms show strong linear relationships in extreme cases (as evidenced by high Pearson correlations), the lower Spearman correlations suggest these relationships are not typical of most humorous content. We observe the following:
\begin{itemize}
    \item Outlier-driven relationships: Particularly between semantic conflicts and rhyme
    \item Stable moderate correlations: Especially in homonym-based features
    \item Independent mechanisms: Such as alliteration and self-reference
    \item Context-dependent interactions: Varying strength of relationships based on humour style and content
\end{itemize}

This analysis suggests that effective humour can be constructed through various combinations of these mechanisms, with no single pattern dominating across all instances.

\subsection{Affective and Emotional Patterns}
Examination of affective patterns revealed distinct emotional signatures and sentiment characteristics across humour styles, providing insights into their underlying psychological mechanisms.

\subsubsection{Distribution of Primary Emotions}
Table \ref{Table:emotion_distribution} presents the frequency distribution of primary emotions across different humour styles. Each style demonstrated characteristic patterns in emotional expression.
\begin{table}[ht!]
\centering
\resizebox{0.8\textwidth}{!}{
    \begin{tabular}{lllllll}
    \hline
    & \multicolumn{6}{c}{\textbf{Emotion Distribution}}    \\ \hline
    & \textbf{Anger} & \textbf{Fear} & \textbf{Joy} & \textbf{Love} & \textbf{Sadness} & \textbf{Surprise} \\ \hline
    \textbf{Affiliative}      & 18             & 8             & 16           & 3             & 3                & 1                 \\ \hline
    \textbf{Aggressive}       & 28             & 7             & 13           & 1             & 8                & 0                 \\ \hline
    \textbf{Neutral}          & 34             & 4             & 39           & 1             & 0                & 2                 \\ \hline
    \textbf{Self-deprecating} & 14             & 4             & 15           & 0             & 11               & 2                 \\ \hline
    \textbf{Self-enhancing}   & 15             & 3             & 30           & 4             & 7                & 2                 \\ \hline
    \end{tabular}
}
\caption{Emotion Distribution by Humour Style}
\label{Table:emotion_distribution}
\end{table}

\paragraph{Self-Enhancing Humour}
Self-enhancing humour demonstrated clear joy (n = 30) predominance with positive sentiment polarity (0.227) and the highest confidence in classification (0.889) among all styles. For instance: \textit{``Self-love is the greatest gift you can give yourself, besides a lifetime supply of cheese''.} This example shows maximum positive sentiment polarity (1.0) and joy (0.978), typical of self-enhancing humour's optimistic orientation. The secondary emotion associated with self-enhancing is anger (n=15) an example of a self-enhancing joke associated with anger is: \textit{``The question isn’t who is going to let me, it’s who is going to stop me”}. This example has anger (0.95) and a positive sentiment (0.99). 

\paragraph{Aggressive Humour}
This showed the highest frequency of anger (n = 28) among all styles, with negative sentiment polarity (-0.039). The presence of joy-related content (n = 13) suggests complex emotional dynamics within this style. Examples of aggressive humour with anger and joy emotions:
\begin{itemize}
    \item \textbf{Anger (0.998):} \textit{``A man wakes from a coma. His wife changes out of her black clothes and, irritated, remarks, “I really cannot depend on you in anything, can I!”}
    \item \textbf{Joy (0.997): }\textit{`` I'd like to leave you with one thought, but I'm not sure you have a place to put it!"}
\end{itemize}
\paragraph{Self-Deprecating Humour}
Analysis revealed that self-deprecating humour maintained the most balanced emotional distribution, with no significant differences between positive and negative emotion frequencies. The distribution showed comparable levels of joy (n = 15, 32.6\%), anger (n = 14, 30.4\%), and sadness (n = 11, 23.9\%), with near-neutral sentiment polarity (mean = -0.002, SD = 0.098).
Examples of self-deprecating humour with anger, joy and sadness emotions:
\begin{itemize}
    \item \textbf{Anger (0.991):} \textit{``You are offered \$50000 but if you accept it, the person you hate most in the entire world gets \$100000. Are you taking it? Yes. Why wouldn't I want \$150000”}
    \item \textbf{Joy (0.997): }\textit{``My memory is as reliable as a goldfish’s. I can forget my own name while introducing myself to someone"}
    \item \textbf{Sadness (0.947):}\textit{ ``I can’t deny that I made a lot of mistakes when I was younger. I’m older now, so I can make different yet more severe mistakes"}
\end{itemize}
\paragraph{Affiliative Humour}
Affiliative humour demonstrated a relatively even distribution between anger (18) and joy (16), with positive overall polarity (0.145), suggesting its role in social bonding through shared emotional experiences.
 Examples of affiliative humour with anger and joy emotions with positive sentiments:
\begin{itemize}
    \item \textbf{Anger ( 0.837): } \textit{``Two cows are in a field. The first one says, “Mooooo!” The second one replies, “that’s what I was going to say!”}
    \item \textbf{Joy (0.549): }\textit{``Money won’t buy happiness, but it will pay the salaries of a large research staff to study the problem"}
\end{itemize} 

\paragraph{Neutral Content} 
Neutral content exhibited high frequencies of both joy (n = 39, 48.8\%) and anger (n = 34, 42.5\%), while maintaining significantly lower subjectivity scores (mean = 0.273, SD = 0.089) compared to all humour styles.  An example of neutral category with joy emotion, positive sentiment and subjectivity of 0.3: \textit{``Buying guide: find the best outdoor patio umbrella for your home"}

\subsubsection{Sentiment and Confidence Patterns}
Table \ref{Table:summary_statistics} presents the distribution of confidence scores and affective metrics across humour styles. Analysis revealed systematic variations in sentiment characteristics and classification confidence across different styles.

\begin{table}[ht!]
\centering
\resizebox{1.0\textwidth}{!}{
    \begin{tabular}{lrrrrr}
\hline
\textbf{Style} & \textbf{N} & \textbf{Confidence} & \textbf{Polarity} & \textbf{Subjectivity} & \textbf{Sarcasm (\%)} \\
\hline
Affiliative & 49 & 0.748 & 0.145 & 0.406 & 18.4 \\
Self-enhancing & 61 & 0.889 & 0.227 & 0.430 & 32.8 \\
Self-deprecating & 46 & 0.811 & -0.002 & 0.474 & 8.7 \\
Aggressive & 57 & 0.780 & -0.039 & 0.404 & 29.8 \\
Neutral & 80 & 0.887 & 0.089 & 0.273 & 48.8 \\
\hline
\end{tabular}
}
\caption{Distribution of Confidence Scores and Affective Metrics Across Humour Styles}
\label{Table:summary_statistics}
\end{table}

\paragraph{Sentiment Polarity}
Analysis revealed a clear demarcation in sentiment polarity across styles. Self-enhancing and affiliative humour maintained positive polarity (0.227 and 0.145 respectively), while aggressive humour exhibited negative polarity (-0.039). Self-deprecating humour demonstrated near-neutral polarity (-0.002), suggesting balanced emotional content. Neutral content showed mild positive polarity (0.089), positioned between the extremes of other styles.

\paragraph{Subjectivity Patterns}
Subjectivity analysis revealed distinct patterns across styles. Self-deprecating humour exhibited the highest subjectivity (0.474), indicating strong emotional investment in content. Conversely, neutral content showed markedly lower subjectivity (0.273), consistent with its more objective nature. Affiliative and aggressive styles showed comparable subjectivity levels (0.406 and 0.404 respectively).

\paragraph{Classification Confidence}
Model confidence demonstrated systematic variation across styles. Self-enhancing humour and neutral content showed the highest confidence scores (0.889 and 0.887 respectively), suggesting more distinctive linguistic and emotional patterns. Affiliative humour exhibited the lowest confidence (0.748), indicating greater classification uncertainty for this style.

\paragraph{Sarcasm Distribution}
Sarcasm presence varied substantially across styles, with neutral content showing the highest frequency (48.8\% of instances) and self-deprecating humour the lowest (8.7\%). This distribution suggests that sarcasm may serve different functions across humour styles, potentially contributing to style differentiation.
Example of sarcastic jokes in the dataset:
\begin{itemize}
    \item \textbf{Self-enhancing:} \textit{“You are never too old to play. You are only too old for low-rise jeans.”}
    \item \textbf{Affiliative: }\textit{“Makeup can only make you look pretty on the outside but it doesn’t help if you’re ugly on the inside. Unless you eat the makeup.”}
    \item \textbf{Self-deprecating:} \textit{“I’m so indecisive, I couldn’t choose my favorite color even if you held a rainbow in front of me!”}
\end{itemize}

These patterns indicate that humour styles employ distinct emotional mechanisms and linguistic strategies. Self-enhancing and affiliative styles maintain predominantly positive emotional valence, while aggressive and self-deprecating styles demonstrate more complex emotional patterns. The variation in subjectivity and sarcasm levels further differentiates these styles, with self-deprecating humour showing heightened emotional investment and neutral content maintaining greater objectivity.
While these patterns reveal distinct characteristics of different humour styles, the classification process faces several challenges, which we examine in detail in the following section on error analysis.

\subsection{Error Analysis and Misclassification Patterns}
Systematic analysis of classification errors revealed distinct patterns in model behaviour, with a predominant challenge in affiliative humour classification. This finding aligns with previously identified performance limitations in the single-model approach \citep{KennethOgbuka2024ARecognition}.

\subsubsection{Primary Misclassification Categories}
Analysis of error patterns revealed three dominant misclassification types, each characterised by distinct feature combinations: 

\paragraph{Affiliative--Neutral Confusion (n = 8)}
These misclassifications demonstrated moderate classification confidence (mean = 0.649) with distinctive sentiment characteristics: elevated polarity (0.197) combined with reduced subjectivity (0.296). In particular, these cases exhibited minimal semantic conflicts (mean = 9.500), suggesting potential oversimplification of humorous content. An example affiliative joke misclassified as neural is: \textit{Joke: ``There are only 3 things that tell the truth: 1 - Young Children 2 - Drunks 3 - Leggings"}.

\paragraph{Affiliative--Self-enhancing Confusion (n = 8)}
Cases in this category showed moderate confidence levels (mean = 0.709) with mild positive sentiment polarity (0.094). A distinguishing feature was the elevated presence of exaggeration (mean = 1.375), indicating potential confusion between social and self-directed humour mechanisms.

\paragraph{Affiliative--Aggressive Confusion (n = 7)}
These cases exhibited the lowest confidence scores (M = 0.611) and negative sentiment polarity (-0.037). The high semantic conflict count (M = 29.857) suggests that complex linguistic structures may contribute to classification ambiguity.\\
Detailed examples of misclassified or confused categories are prsented in subsection 4.4.3. 

\subsubsection{Feature-Specific Error Analysis}
Analysis of classification errors revealed systematic patterns across multiple dimensions, as detailed in Table \ref{Table:error_characteristics}.

\paragraph{Confidence Patterns} 
Model confidence demonstrated significant variation across error types (p \textless 0.0001), with distinct patterns emerging:
\begin{itemize}
\item Affiliative--aggressive misclassifications exhibited the lowest confidence (mean = 0.611, SD = 0.179).
\item Self-deprecating misclassifications showed unexpectedly high confidence (mean = 0.811, SD = 0.160).
\end{itemize}

\paragraph{Sentiment Characteristics}
Sentiment analysis revealed significant associations between misclassification types and polarity patterns (p = 0.0001):
\begin{itemize}
\item Affiliative humour misclassified as aggressive showed negative polarity (-0.037), deviating from typical affiliative patterns.
\item Self-deprecating humour misclassified as affiliative exhibited unexpected positive sentiment (0.403).
\item Subjectivity scores varied systematically with error types, particularly in cases involving self-deprecating humour.
\end{itemize}

\paragraph{Semantic Structure}
Semantic conflict analysis revealed significant patterns (p = 0.0003):
\begin{itemize}
    \item The presence of high semantic conflict counts (\textgreater 25) strongly influenced misclassifications.
    \item Affiliative--aggressive confusions showed elevated semantic conflicts (mean = 29.857, SD = 63.006).
    \item Aggressive--affiliative misclassifications demonstrated the highest semantic conflict counts (mean = 37.833, SD = 74.778).
\end{itemize}

\begin{table}[ht!]
\centering
\small
\begin{tabular}{l|rr|rr|rr}
\hline
\multirow{2}{*}{\textbf{Error Type}} & \multicolumn{2}{c|}{\textbf{Confidence}} & \multicolumn{2}{c|}{\textbf{Semantic Conflicts}} & \multicolumn{2}{c}{\textbf{Sentiment}} \\
& Mean & SD & Mean & SD & Polarity & Subj. \\
\hline
\multicolumn{7}{l}{\textit{Affiliative Misclassifications}} \\
\hline
→ Aggressive & 0.611 & 0.179 & 29.857 & 63.006 & -0.037 & 0.304 \\
→ Neutral & 0.649 & 0.226 & 9.500 & 7.309 & 0.197 & 0.296 \\
→ Self-deprecating & 0.776 & 0.289 & 22.000 & 15.620 & 0.175 & 0.510 \\
→ Self-enhancing & 0.709 & 0.169 & 6.500 & 6.908 & 0.094 & 0.256 \\
\hline
\multicolumn{7}{l}{\textit{Aggressive Misclassifications}} \\
\hline
→ Affiliative & 0.663 & 0.178 & 37.833 & 74.778 & -0.024 & 0.568 \\
→ Neutral & 0.746 & 0.171 & 7.600 & 8.792 & 0.016 & 0.150 \\
→ Self-deprecating & 0.628 & 0.227 & 10.500 & 9.192 & 0.300 & 0.442 \\
→ Self-enhancing & 0.698 & 0.268 & 0.500 & 0.707 & 0.356 & 0.750 \\
\hline
\multicolumn{7}{l}{\textit{Self-deprecating Misclassifications}} \\
\hline
→ Affiliative & 0.775 & 0.074 & 17.333 & 11.015 & 0.403 & 0.550 \\
→ Aggressive & 0.811 & 0.160 & 14.000 & 7.937 & -0.046 & 0.554 \\
→ Neutral & 0.773 & 0.159 & 5.500 & 7.778 & 0.000 & 0.275 \\
→ Self-enhancing & 0.881 & -- & 28.000 & -- & 0.250 & 0.250 \\
\hline
\multicolumn{7}{l}{\textit{Self-enhancing Misclassifications}} \\
\hline
→ Affiliative & 0.587 & 0.423 & 7.000 & 4.243 & -0.250 & 0.500 \\
→ Aggressive & 0.745 & -- & 0.000 & -- & 0.062 & 0.167 \\
→ Neutral & 0.480 & 0.026 & 11.500 & 2.121 & 0.000 & 0.000 \\
→ Self-deprecating & 0.688 & 0.264 & 6.800 & 7.791 & -0.076 & 0.383 \\
\hline
\end{tabular}
\caption{Characteristics of Misclassification Types}
\label{Table:error_characteristics}
\end{table}

\paragraph{Additional Feature Patterns} Analysis of supplementary features revealed:
\begin{itemize}
     \item Confidence scores generally decreased when the model confused similar styles (e.g, affiliative and aggressive).
    \item Sentiment polarity often deviated significantly from expected patterns for specific humour styles.
    \item Higher semantic conflict counts corresponded with increased likelihood of misclassification between affiliative and aggressive styles.
    \item Sarcasm probability showed significant variation across error types, suggesting its role in misclassifications.
\end{itemize}

These patterns indicate that misclassifications arise from complex interactions between linguistic and affective features, particularly when these features deviate from style-typical patterns. The systematic nature of these deviations suggests specific weaknesses in the model's ability to distinguish between stylistically similar humour categories, especially in cases with high semantic complexity.

\subsubsection{Analysis of Representative Misclassifications}
Detailed examination of misclassified instances revealed systematic patterns in classification errors. We selected eight representative cases (two from each humour style) to illustrate key error patterns, focusing on cases with the most distinctive feature combinations.

\paragraph{Affiliative Humour Misclassifications}
\begin{enumerate}
    \item \textbf{Error Type: Affiliative misclassified as Self-deprecating:}
    \begin{itemize}
        \item \textit{Joke: ``As best man, it is my job to tell you about the groom, and all the embarrassing things that have happened to him in the 28 years leading up to what was the happiest day of his life until I started this speech"}
        \item confidence score: 0.447
        \item Top features: {``embarrassing": 0.30, ``my": 0.17, ``groom": -0.17, ``i": 0.12, ``28": -0.10}
        \item Linguistic features: High semantic conflict count (40), self-reference count (2 instances) 
        \item Affective Profile: Sarcasm (false), negative sentiment (0.989 confidence), subjectivity (0.3), emotion (sadness—0.932 confidence) 
        \item Target: {self-targeted: true, other-targeted: true, situation-targeted: false}
        \item Analysis: The model focusing on the word "embarrassing" and personal references  "my", and the strong negative sentiment and emotion, and high semantic conflict in the sentence triggered self-deprecating classification despite the social bonding context typical of wedding speeches. 
    \end{itemize}
    \item \textbf{Error Type: Affiliative misclassified as Aggressive:}
    \begin{itemize}
        \item \textit{Joke: ``What did one DNA say to the other DNA? these genes make me look fat}"
        \item confidence score: 0.781
        \item Top features: {``me": 0.05, ``say": -0.03, ``genes": -0.03, ``one": -0.02, ``did": -0.02}
        \item Linguistic features: semantic conflicts (19), self-reference count (1) 
        \item Affective Profile: Sarcasm probability (0.999), negative sentiment (0.99 confidence), subjectivity (0.375), emotion (joy—0.865 confidence)
        \item Target: {self-targeted: true, other-targeted: false, situation-targeted: false}
        \item Analysis: The model's high confidence (0.781) in the aggressive classification appears driven by the sentence's sarcastic nature, negative sentiment, and homophones ``genes/jeans" that could be interpreted as mockery. 
    \end{itemize} 
\end{enumerate}

\paragraph{Aggressive Humour Misclassifications}
\begin{enumerate}
    \item \textbf{Error Type: Aggressive misclassified as Affiliative:} 
    \begin{itemize}
        \item \textit{Joke: ``After every sentence I say you say ketchup and rubber buns. what did you eat for breakfast? ``ketchup \& rubber buns." what did you eat for lunch? ``ketchup \& rubber buns." what did you eat for dinner? ``ketchup \& rubber buns." what do you do when you see a hot girl? ``ketchup \& rubber buns." YOU WHERE RUBBING MY GF'S WHAT?!?!"}
        \item confidence score: 0.42
        \item Top features: {``every": 0.08, ``I": 0.06, ``say": -0.04, ``sentence": 0.03, ``MY": 0.03}
        \item Linguistic features: semantic conflicts (190), self-reference count (2), rhyme count (14)
        \item Affective Profile: Sarcasm (false), positive sentiment (0.99 confidence), subjectivity (0.85), emotion (love - 0.98 confidence)
        \item Target: {self-targeted: true, other-targeted: true, situation-targeted: false}
        \item Analysis: The model's misclassification appears driven by the high positive sentiment, high positive emotion of love, and playful repetition (high rhyme count) rather than the hostile twist ending. 
    \end{itemize}
    
    \item \textbf{Error Type: Aggressive misclassified as Self-enhancing:}
    \begin{itemize}
        \item \textit{Joke: Cats have nine lives. Makes them ideal for experimentation}
        \item confidence score: 0.51
        \item Top features: {"experimentation": 0.03, "nine": -0.02, "have": -0.01, "Cats": -0.009, "Makes": -0.009}
        \item Linguistic features: semantic conflicts (1), self-reference count (2)
        \item Affective Profile: Sarcasm probability (1.0), positive sentiment (0.99 confidence), subjectivity (1.0), emotion (joy - 0.99 confidence)
        \item Target: {self-targeted: false, other-targeted: false, situation-targeted: true}
        \item Analysis: The model's misclassification appears driven by the high positive sentiment, high positive emotion of joy, presence of sarcasm, and lack of typical aggressive markers like profanity or direct attacks. 
    \end{itemize}
    
\end{enumerate}

\paragraph{Self-deprecating Humour Misclassifications}
\begin{enumerate}
    \item \textbf{Error Type: Self-deprecating misclassified as Self-enhancing:}  
    \begin{itemize}
        \item \textit{Joke: ``You're guessing that out of the 8 billion people here on Earth, I'm going to chase someone who doesn't even like me? Well, watch me closely, because that's exactly what I'm going to do"} 
        \item confidence score: 0.88
        \item Top features: {``me": 0.12, ``Well": -0.09, ``chase": -0.08, ``I": 0.08, ``You": -0.07}
        \item Linguistic features: semantic conflicts (28), self-reference count (4)
        \item Affective Profile: Sarcasm (false), negative sentiment (0.99 confidence), subjectivity (0.25), emotion (joy - 0.65 confidence)
        \item Target: {self-targeted: true, other-targeted: true, situation-targeted: false}
        \item Analysis: The high self-reference makes the joke to be likely classified as either self-deprecating or self-enhancing. Even though the sentiment was negative, the misclassification arises from the emotion being joy, which conflicts with the sentiment and the high semantic conflicts of 28 reflect this.
    \end{itemize}

    \item \textbf{Error Type: Self-deprecating misclassified as Aggressive:}
    \begin{itemize}
        \item \textit{Joke: ``What would have happened if you exterminated the ugliest guy and the dumbest guy in the world yesterday? Right, this post wouldn't exist."} 
        \item confidence score: 0.99
        \item Top features: {``dumbest": -0.02, ``guy": 0.01, ``would": -0.008, ``wouldn't": -0.005, ``Right": 0.005}
        \item Linguistic features: semantic conflicts (20), self-reference count (0)
        \item Affective Profile: Sarcasm probability (1.0), negative sentiment (0.99 confidence), subjectivity (0.55), emotion (anger - 0.65 confidence)
        \item Target: {self-targeted: false, other-targeted: true, situation-targeted: false}
        \item Analysis: Negative sentiment and emotions are both highly correlated with self-deprecating and aggressive humour; however, the lack of self-reference and the presence of others-reference led the model to mislabel.  
    \end{itemize}
\end{enumerate}

\paragraph{Self-enhancing Humour Misclassifications}
\begin{enumerate}
    \item \textbf{Error Type: Self-enhancing misclassified as Affiliative:}
    \begin{itemize}
        \item \textit{Jokes: ``I was gonna tell a joke about pizza, but it's too cheesy"}
        \item confidence score: 0.29
        \item Top features: {``I": 0.21, ``pizza": -0.18, ``joke": -0.14, ``cheesy": 0.12, ``tell": -0.09}
        \item Linguistic features: semantic conflicts (10), self-reference count (1)
        \item Affective Profile: Sarcasm (false), negative sentiment (0.99 confidence), subjectivity (1.0), emotion (joy - 0.90 confidence)
        \item Target: {self-targeted: true, other-targeted: false, situation-targeted: false}
        \item Analysis: The low confidence score (0.29) reflects uncertainty, indicating model difficulty. The self-referential context (``I") and playful tone suggest self-enhancing humour. However, the light-hearted nature, lack of personal boasting, and emphasis on social food humour, coupled with high joy emotion (0.90), likely caused misclassification as affiliative humour.
    \end{itemize}

    \item\textbf{Error Type: Self-enhancing misclassified as Self-deprecating:} 
    \begin{itemize}
        \item \textit{Joke: ``I went to the doctor with a strawberry growing on my bum. The doctor said I've got some cream for that"}
        \item confidence score: 0.33
        \item Top features: {``strawberry": -0.20, ``I": 0.19, ``growing": -0.16, ``bum": 0.10, ``said": -0.10}
        \item Linguistic features: semantic conflicts (4), self-reference count (3)
        \item Affective Profile: Sarcasm (false), negative sentiment (0.99 confidence), subjectivity (0.0), emotion (anger - 0.54 confidence)
        \item Target: {self-targeted: true, other-targeted: false, situation-targeted: false}
        \item Analysis: The misclassification probably resulted from the overlap of humour characteristics. The joke’s self-referential nature (3 instances) and mildly embarrassing scenario align with self-deprecating humour. However, the speaker's light-hearted tone and ability to laugh at a personal situation suggest self-enhancing intent. Low semantic conflict (4) and the unexpected association of "anger" as the dominant emotion further confused the model. 
    \end{itemize}
\end{enumerate}
\textbf{}\\
\textbf{Error Pattern Analysis}\\
Analysis of the misclassified instances revealed five fundamental challenges in humour style classification:

\begin{enumerate}
    \item \textbf{Emotional Ambiguity:} The model demonstrates significant sensitivity to conflicting emotional signals. This is particularly evident in self-enhancing humour, where positive sentiment often co-occurs with contradictory emotional markers (e.g., anger or fear), leading to misalignment between detected and intended emotional content.
    
    \item \textbf{Context Misinterpretation:} The model shows systematic difficulties in accurately interpreting social contexts, particularly in discriminating between self-directed and other-directed humorous content. This limitation suggests insufficient incorporation of broader contextual cues in the classification process.
    
    \item \textbf{Self-Reference Confusion:} Self-referential content creates particular classification challenges, most notably in distinguishing between self-deprecating and self-enhancing styles. This suggests potential limitations in the model's ability to interpret the valence of self-referential statements.
    
    \item \textbf{Target Ambiguity:} The model exhibits reduced accuracy in cases with multiple potential targets (self, others, or situations). This limitation is particularly pronounced in instances where humorous content contains overlapping targeting mechanisms.
    
    \item \textbf{Feature Interference:} The simultaneous presence of multiple humour mechanisms (e.g., wordplay, sarcasm, semantic conflicts) creates competing classification signals, suggesting limitations in the model's ability to integrate multiple linguistic features.
\end{enumerate}

These patterns demonstrate the effectiveness of the XAI framework in identifying both feature-level drivers of classification decisions and systematic classification challenges. The findings suggest that improving model performance will require enhanced handling of cases involving multiple interacting humour mechanisms, particularly for affiliative humour where the interplay between emotional and linguistic signals exhibits greater complexity.

\section{Conclusion and Future Work}
This study introduces a comprehensive XAI framework for understanding humour style classification, revealing significant patterns in how linguistic mechanisms and emotional features interact across humour styles. Social forms of humour, particularly affiliative and aggressive styles, demonstrate higher rates of wordplay and semantic conflicts. The emotional signatures vary markedly between styles, with self-enhancing humour showing consistently positive polarity (0.227) and high confidence (0.889), while aggressive humour tends toward negative polarity (-0.039). Our correlation analyses reveal complex relationships between linguistic mechanisms, particularly in the relationship between semantic conflicts and rhyme, where Pearson correlation (r = 0.95) suggests a strong linear relationship, but the much lower Spearman correlation (r = 0.32) indicates this relationship is heavily influenced by outliers rather than being typical across most instances.

The classification challenges highlight important considerations for future development. The model shows varying confidence levels across styles, particularly struggling with affiliative humour (lowest average confidence: 0.748). These challenges often stem from emotional ambiguity and context misinterpretation, especially when multiple humour mechanisms interact. Self-reference and target identification emerged as critical factors in classification decisions, suggesting their importance for future model development.

The practical implications of our findings extend beyond theoretical understanding. Our framework provides actionable insights for researchers studying humour's role in psychological well-being and social communication, while offering a valuable template for applying XAI to other subjective classification tasks in digital humanities. The framework's ability to explain model decisions enhances transparency and trust in automated humour analysis, supporting applications in mental health assessment, content moderation, and digital humanities research.

Despite the valuable insights provided by this study, several important limitations must be acknowledged. Our analysis focuses on a single model configuration (ALI+XGBoost) with a dataset of 1,463 English-language instances, which inherently limits the generalisability of our findings. The feature detection components faced significant technical constraints, particularly in terms of dictionary coverage and algorithmic accuracy. The CMU Pronouncing Dictionary covered 90.75\% of unique words in our dataset for rhyme detection, while WordNet synset coverage reached 88.50\% of the vocabulary, leaving significant gaps in handling slang and neologisms. The alliteration detector showed particular limitations, such as considering only the first phoneme, leading to false positives like grouping words beginning with different sounds (e.g.,  ``when/one"), counting repeated words as alliteration, and grouping words regardless of their proximity in the text. This contributed to moderate correlations with homonym detection (Pearson: 0.21, Spearman: 0.27), which is partially due to overlapping feature detection rather than true linguistic relationships.

A significant methodological constraint lies in our reliance on pre-trained models for sarcasm (60.7\% accuracy), sentiment (93.2\% accuracy), and emotion detection (93.8\% accuracy), not specifically trained on humorous content. Humorous text contains unique linguistic patterns, irony, and complex emotional layers that may not be well-captured by these general-purpose models. The binary classification of features, such as sarcasm detection, oversimplifies complex linguistic phenomena that exist on a spectrum. Additionally, automated sentiment and emotion detection may not fully capture the subtle, often contradictory emotional cues present in different humour styles, like simultaneous positive and negative emotions in self-deprecating humour.

Several analytical constraints affect our findings. The study does not address temporal or contextual variations in humour interpretation, and the analysis of misclassifications may not capture all possible error patterns due to the limited sample size. Future research should expand to multiple models and larger, more diverse datasets, incorporating cross-cultural perspectives and multilingual analysis. Enhanced feature detection algorithms are needed, particularly for complex linguistic features like multi-word puns and contextual wordplay, along with better handling of informal language. The development of specialised models trained specifically on humorous content would improve the accuracy of sentiment and emotion detection in comedic contexts. Additionally, more refined methods for capturing contextual and cultural factors would enhance the framework's applicability across different contexts, while integration of temporal analysis would provide deeper insights into how humour interpretation varies over time and across different social settings. These improvements would strengthen the framework's reliability and broaden its applicability in real-world scenarios.

\section{Acknowledgement}
This research was supported by the Petroleum Technology Development Fund (PTDF) of Nigeria.

\bibliographystyle{plainnat}
\bibliography{references}

\begin{thebibliography}{31}
\providecommand{\natexlab}[1]{#1}
\providecommand{\url}[1]{\texttt{#1}}
\expandafter\ifx\csname urlstyle\endcsname\relax
  \providecommand{\doi}[1]{doi: #1}\else
  \providecommand{\doi}{doi: \begingroup \urlstyle{rm}\Url}\fi

\bibitem[Abulaish and Kamal(2018)]{Abulaish2018Self-DeprecatingApproach}
Muhammad Abulaish and Ashraf Kamal.
\newblock {Self-Deprecating Sarcasm Detection: An Amalgamation of Rule-Based and Machine Learning Approach}.
\newblock In \emph{Proceedings - 2018 IEEE/WIC/ACM International Conference on Web Intelligence, WI 2018}, pages 574--579. Institute of Electrical and Electronics Engineers Inc., 1 2018.
\newblock ISBN 9781538673256.
\newblock \doi{10.1109/WI.2018.00-35}.

\bibitem[Ahmed et~al.(2022)Ahmed, Jhaveri, Srivastava, and Lin]{Ahmed2022ExplainableDisorder}
Usman Ahmed, Rutvij~H. Jhaveri, Gautam Srivastava, and Jerry Chun-Wei Lin.
\newblock {Explainable Deep Attention Active Learning for Sentimental Analytics of Mental Disorder}.
\newblock \emph{ACM Transactions on Asian and Low-Resource Language Information Processing}, 7 2022.
\newblock ISSN 2375-4699.
\newblock \doi{10.1145/3551890}.

\bibitem[Amjad and Dasti(2022)]{Amjad2022HumorAdults}
Arooba Amjad and Rabia Dasti.
\newblock {Humor styles, emotion regulation and subjective well-being in young adults}.
\newblock \emph{Current Psychology}, 41\penalty0 (9):\penalty0 6326--6335, 9 2022.
\newblock ISSN 19364733.
\newblock \doi{10.1007/s12144-020-01127-y}.

\bibitem[Anderson and Di~Tunnariello(2016)]{Anderson2016AggressiveConflict}
Whitney Anderson and Nancy Di~Tunnariello.
\newblock {Aggressive humor as a negative relational maintenance behavior during times of conflict}.
\newblock \emph{Qualitative Report}, 21\penalty0 (8):\penalty0 1513--1530, 8 2016.
\newblock ISSN 21603715.
\newblock \doi{10.46743/2160-3715/2016.2149}.

\bibitem[Annamoradnejad and Zoghi(2020)]{Annamoradnejad2020ColBERT:Humor}
Issa Annamoradnejad and Gohar Zoghi.
\newblock {ColBERT: Using BERT Sentence Embedding in Parallel Neural Networks for Computational Humor}.
\newblock 4 2020.
\newblock URL \url{http://arxiv.org/abs/2004.12765}.

\bibitem[Chen and Martin(2007)]{Chen2007AStudents}
Guo~Hai Chen and Rod~A. Martin.
\newblock {A comparison of humor styles, coping humor, and mental health between Chinese and Canadian university students}.
\newblock \emph{Humor}, 20\penalty0 (3):\penalty0 215--234, 8 2007.
\newblock ISSN 09331719.
\newblock \doi{10.1515/HUMOR.2007.011}.

\bibitem[Chen et~al.(2024)Chen, Yuan, Liu, Liu, Guan, Guo, Peng, Liu, Li, and Xiao]{Chen2024TalkInterpretation}
Yuyan Chen, Yichen Yuan, Panjun Liu, Dayiheng Liu, Qinghao Guan, Mengfei Guo, Haiming Peng, Bang Liu, Zhixu Li, and Yanghua Xiao.
\newblock {Talk Funny! A Large-Scale Humor Response Dataset with Chain-of-Humor Interpretation}.
\newblock In \emph{The Thirty-Eighth AAAI Conference on Artiﬁcial Intelligence (AAAI-24)}, pages 17826--17834, 2024.
\newblock URL \url{https://ojs.aaai.org/index.php/AAAI/article/view/29736}.

\bibitem[Chowdhury et~al.(2021)Chowdhury, Sil, and Shukla]{Chowdhury2021ExplainingLIME}
Kounteyo~Roy Chowdhury, Arpan Sil, and Sharvari~Rahul Shukla.
\newblock {Explaining a Black-Box Sentiment Analysis Model with Local Interpretable Model Diagnostics Explanation (LIME)}.
\newblock In \emph{Communications in Computer and Information Science}, volume 1440 CCIS, pages 90--101. Springer Science and Business Media Deutschland GmbH, 2021.
\newblock ISBN 9783030814618.
\newblock \doi{10.1007/978-3-030-81462-5{\_}9}.

\bibitem[Cortinas-Lorenzo and Lacey(2024)]{Cortinas-Lorenzo2024TowardReview}
Karina Cortinas-Lorenzo and Gerard Lacey.
\newblock {Toward Explainable Affective Computing: A Review}.
\newblock \emph{IEEE Transactions on Neural Networks and Learning Systems}, 2024.
\newblock ISSN 21622388.
\newblock \doi{10.1109/TNNLS.2023.3270027}.

\bibitem[De~Marez et~al.(2024)De~Marez, Winters, and Terryn]{DeMarez2024THInC:Detection}
Victor De~Marez, Thomas Winters, and Ayla~Rigouts Terryn.
\newblock {THInC: A Theory-Driven Framework for Computational Humor Detection}.
\newblock In \emph{InternationalWorkshop on Artificial Intelligence and Creativity}, Santiago de Compostela,, 9 2024.
\newblock URL \url{http://arxiv.org/abs/2409.01232}.

\bibitem[Edalat(2023)]{Edalat2023Self-initiatedLaugh}
Abbas Edalat.
\newblock {Self-initiated humour protocols: An algorithmic approach for learning to laugh}.
\newblock \emph{PsyArXiv}, 5:\penalty0 1--14, 11 2023.
\newblock \doi{https://doi.org/10.31234/osf.io/w9cvx}.

\bibitem[Edalat et~al.(2024)Edalat, Hu, Patel, Polydorou, Ryan, and Nicholls]{Edalat2024Self-InitiatedAgent}
Abbas Edalat, Ruoyu Hu, Zeena Patel, Neophytos Polydorou, Frank Ryan, and Dasha Nicholls.
\newblock {Self-Initiated Humour Protocol: A pilot study with an AI agent}.
\newblock \emph{PsyArXiv}, 2024.
\newblock \doi{10.31234/osf.io/f32xq}.

\bibitem[Hartmann et~al.(2023)Hartmann, Heitmann, Siebert, and Schamp]{Hartmann2023MoreAnalysis}
Jochen Hartmann, Mark Heitmann, Christian Siebert, and Christina Schamp.
\newblock {More than a Feeling: Accuracy and Application of Sentiment Analysis}.
\newblock \emph{International Journal of Research in Marketing}, 40\penalty0 (1):\penalty0 75--87, 3 2023.
\newblock ISSN 01678116.
\newblock \doi{10.1016/j.ijresmar.2022.05.005}.

\bibitem[Kamal and Abulaish(2020)]{Kamal2020Self-deprecatingApproach}
Ashraf Kamal and Muhammad Abulaish.
\newblock {Self-deprecating Humor Detection: A Machine Learning Approach}.
\newblock In Le-Minh Nguyen, Xuan-Hieu Phan, Kôiti Hasida, and Satoshi Tojo, editors, \emph{Computer Lingustics}, volume 1215 of \emph{Communications in Computer and Information Science}, pages 483--484. Springer Singapore, Singapore, 2020.
\newblock ISBN 978-981-15-6167-2.
\newblock \doi{10.1007/978-981-15-6168-9}.
\newblock URL \url{http://link.springer.com/10.1007/978-981-15-6168-9}.

\bibitem[Kazienko et~al.(2023)Kazienko, Bielaniewicz, Gruza, Kanclerz, Karanowski, Mi{\l}kowski, and Koco{\'{n}}]{Kazienko2023Human-centeredHumor}
Przemysław Kazienko, Julita Bielaniewicz, Marcin Gruza, Kamil Kanclerz, Konrad Karanowski, Piotr Mi{\l}kowski, and Jan Koco{\'{n}}.
\newblock {Human-centered neural reasoning for subjective content processing: Hate speech, emotions, and humor}.
\newblock \emph{Information Fusion}, 94:\penalty0 43--65, 6 2023.
\newblock ISSN 15662535.
\newblock \doi{10.1016/j.inffus.2023.01.010}.

\bibitem[Kenneth et~al.(2024)Kenneth, Khosmood, and Edalat]{Kenneth2024SystematicClassification}
Mary~Ogbuka Kenneth, Foaad Khosmood, and Abbas Edalat.
\newblock {Systematic Literature Review: Computational Approaches for Humour Style Classification}.
\newblock Technical report, 2024.
\newblock URL \url{https://www.semanticscholar.}

\bibitem[Kenneth~Ogbuka et~al.(2024)Kenneth~Ogbuka, Khosmood, and Edalat]{KennethOgbuka2024ARecognition}
Mary Kenneth~Ogbuka, Foaad Khosmood, and Abbas Edalat.
\newblock {A Two-Model Approach for Humour Style Recognition}.
\newblock In \emph{Proceedings of the 4th International Conference on Natural Language Processing for Digital Humanities}, pages 259--274, Miami, 11 2024. Association for Computational Linguistics.
\newblock URL \url{https://aclanthology.org/2024.nlp4dh-1.25/}.

\bibitem[Kuiper et~al.(2016)Kuiper, Kirsh, and Maiolino]{Kuiper2016IdentityWell-Being}
Nicholas Kuiper, Gillian Kirsh, and Nadia Maiolino.
\newblock {Identity and Intimacy Development, Humor Styles, and Psychological Well-Being}.
\newblock \emph{Identity}, 16\penalty0 (2):\penalty0 115--125, 4 2016.
\newblock ISSN 1532706X.
\newblock \doi{10.1080/15283488.2016.1159964}.

\bibitem[Lundberg and Lee(2017)]{Lundberg2017APredictions}
Scott Lundberg and Su-In Lee.
\newblock {A Unified Approach to Interpreting Model Predictions}.
\newblock In \emph{31st Conference on Neural Information Processing Systems}, Long Beach, 5 2017.
\newblock URL \url{http://arxiv.org/abs/1705.07874}.

\bibitem[Lyu et~al.(2024)Lyu, Apidianaki, and Callison-Burch]{Lyu2024TowardsSurvey}
Qing Lyu, Marianna Apidianaki, and Chris Callison-Burch.
\newblock {Towards Faithful Model Explanation in NLP: A Survey}.
\newblock \emph{Computational Linguistics}, pages 657--723, 2024.
\newblock \doi{10.1162/coli}.
\newblock URL \url{https://doi.org/10.1162/coli}.

\bibitem[Mahajan and Zaveri(2024)]{Mahajan2024AnModels}
Rutal Mahajan and Mukesh Zaveri.
\newblock {An automatic humor identification model with novel features from Berger's typology and ensemble models}.
\newblock \emph{Decision Analytics Journal}, 11, 6 2024.
\newblock ISSN 27726622.
\newblock \doi{10.1016/j.dajour.2024.100450}.

\bibitem[Mann and Mikulandric(2024)]{Mann2024CLEFClassification}
Rowan Mann and Tomislav Mikulandric.
\newblock {CLEF 2024 JOKER Tasks 1-3: Humour identification and classification}.
\newblock In \emph{Conference and Labs of the Evaluation Forum}, Grenoble, 9 2024.

\bibitem[Martin and Ford(2018)]{Martin2018TheApproach}
Rod~A Martin and Thomas Ford.
\newblock \emph{{The psychology of humor: An integrative approach}}.
\newblock Academic Press, 2nd edition, 2018.

\bibitem[Martin et~al.(2003)Martin, Puhlik-Doris, Larsen, Gray, and Weir]{Martin2003IndividualQuestionnaire}
Rod~A Martin, Patricia Puhlik-Doris, Gwen Larsen, Jeanette Gray, and Kelly Weir.
\newblock {Individual differences in uses of humor and their relation to psychological well-being: Development of the Humor Styles Questionnaire}.
\newblock \emph{Journal of Research in Personality}, 37:\penalty0 48--75, 2003.
\newblock URL \url{www.elsevier.com/locate/jrp}.

\bibitem[Ortega-Bueno et~al.(2022)Ortega-Bueno, Rosso, and Medina~Pagola]{Ortega-Bueno2022Multi-viewVariants}
Reynier Ortega-Bueno, Paolo Rosso, and José~E. Medina~Pagola.
\newblock {Multi-view informed attention-based model for Irony and Satire detection in Spanish variants}.
\newblock \emph{Knowledge-Based Systems}, 235, 1 2022.
\newblock ISSN 09507051.
\newblock \doi{10.1016/j.knosys.2021.107597}.

\bibitem[P{\'{e}}rez-Landa et~al.(2021)P{\'{e}}rez-Landa, Loyola-Gonz{\'{a}}lez, and Medina-P{\'{e}}rez]{Perez-Landa2021AnTweets}
Gabriel~Ichcanziho P{\'{e}}rez-Landa, Octavio Loyola-Gonz{\'{a}}lez, and Miguel~Angel Medina-P{\'{e}}rez.
\newblock {An explainable artificial intelligence model for detecting xenophobic tweets}.
\newblock \emph{Applied Sciences (Switzerland)}, 11\penalty0 (22), 11 2021.
\newblock ISSN 20763417.
\newblock \doi{10.3390/app112210801}.

\bibitem[Ribeiro et~al.(2016)Ribeiro, Singh, and Guestrin]{Ribeiro2016WhyClassifier}
Marco~Tulio Ribeiro, Sameer Singh, and Carlos Guestrin.
\newblock {"Why Should I Trust You?" Explaining the Predictions of Any Classifier}.
\newblock In \emph{Proceedings of NAACL-HLT}, pages 97--101, San Diego, 6 2016. Association for Computational Linguistics.
\newblock URL \url{https://github.}

\bibitem[Sari(2016)]{Sari2016WasHumor}
Serkan~Volkan Sari.
\newblock {Was it just joke? Cyberbullying perpetrations and their styles of humor}.
\newblock \emph{Computers in Human Behavior}, 54:\penalty0 555--559, 1 2016.
\newblock ISSN 07475632.
\newblock \doi{10.1016/j.chb.2015.08.053}.

\bibitem[Sen(2012)]{Sen2012HumourResearch}
Anindya Sen.
\newblock {Humour Analysis and Qualitative Research}.
\newblock Technical report, University of Surrey, 2012.

\bibitem[Zhang et~al.(2017)Zhang, Song, Liu, Du, and Zhao]{Zhang2017InvestigationsRecognition}
Donghai Zhang, Wei Song, Lizhen Liu, Chao Du, and Xinlei Zhao.
\newblock {Investigations in automatic humor recognition}.
\newblock In \emph{Proceedings - 2017 10th International Symposium on Computational Intelligence and Design, ISCID 2017}, volume 1 2018-January, pages 272--275. Institute of Electrical and Electronics Engineers Inc., 7 2017.
\newblock ISBN 9781538636749.
\newblock \doi{10.1109/ISCID.2017.160}.

\bibitem[Zhu et~al.(2022)Zhu, Ou, and Zhu]{Zhu2022AggressivePerspective}
Hong Zhu, Yilin Ou, and Zimeng Zhu.
\newblock {Aggressive humor style and cyberbullying perpetration: Normative tolerance and moral disengagement perspective}.
\newblock \emph{Frontiers in Psychology}, 13, 12 2022.
\newblock ISSN 16641078.
\newblock \doi{10.3389/fpsyg.2022.1095318}.

\end{thebibliography}

\appendix\footnotesize
\section{Summary of Related works}
\label{appendix:related_works}
Table \ref{Table:related_works} summarises the key works discussed in section 2 (Related works), highlighting the authors' specific tasks, datasets used, features extracted, classification approaches, XAI methods employed, and their achieved results. This overview demonstrates both the progress made in computational humour analysis and explainability techniques while highlighting the gaps our work addresses.

\begin{table}[ht!]
\centering
\resizebox{1.0\textwidth}{!}{
    \begin{tabular}{lllllll}
    \hline
    \textbf{Author}            & \textbf{Task} & \textbf{Dataset}  & \textbf{Extracted Features}    & \textbf{Classifiers}                & \textbf{XAI Method}  & \textbf{Results}   \\ \hline
    
    \citet{Perez-Landa2021AnTweets} & Xenophobia    & \begin{tabular}[c]{@{}l@{}}Experts Xenophobia\\ Database (EXD) (10,057 tweets)\\ Pitropakis Xenophobia Database\\ (PXD)(5814 tweets)\end{tabular}  & \begin{tabular}[c]{@{}l@{}}TF-IDF, Bag of words (BOW),\\ Word 2 vector (W2V),  \\ Keywords, Sentiment, \\ Emotion, and Syntactic\end{tabular}     & \begin{tabular}[c]{@{}l@{}}C45, KNN, Rusboost,\\ UnderBagging,\end{tabular}                                & \begin{tabular}[c]{@{}l@{}}Feature engineering\\ (keywords, sentiment,\\ emotions and syntactic\end{tabular}   & \begin{tabular}[c]{@{}l@{}}F1: 76.8\% (EXD)\\ AUC: 0.864 (EXD)\\ F1: 73.4\% (PXD)\\ AUC: 0.794 (PXD)\end{tabular}     \\ \hline
    
     \cite{Chowdhury2021ExplainingLIME}  & Sentiment     & Twitter API (2-years period)  & \begin{tabular}[c]{@{}l@{}}W2V, Sentiment Speciﬁc Word \\ Embedding (SSWE), GloVe,\\ FastText embeddings\end{tabular}    & Bi-directional LSTM     & LIME    & \begin{tabular}[c]{@{}l@{}}ACC: 72\%\\ F1: 72\%\end{tabular}  \\ \hline
    
    \cite{Ahmed2022ExplainableDisorder}& Sentiment     & Patient-authored text (15,044) 
    & \begin{tabular}[c]{@{}l@{}}Fuzzy logic rules, GloVe, \\ Cosine Similarity, Keywords\end{tabular}  & \begin{tabular}[c]{@{}l@{}}Bi-directional LSTM \\ Attention with fuzzy\\ classification\end{tabular}     & Fuzzy logic rules   & F1: 89\%    \\ \hline
    
    \cite{Ortega-Bueno2022Multi-viewVariants}       & Irony, Satire & \begin{tabular}[c]{@{}l@{}}IroSvA’19 shared task (3000)\\ Barbieri’15-es (10,000)\\ Salas’17-mx (10,888)\\ HAHA’19 shared task (6000)\end{tabular} & \begin{tabular}[c]{@{}l@{}}BERT embedding, \\ Multilingual Universal \\ Sentence encoding, \\ semantics, affective, incongruity,\\ stylistic and structural\end{tabular} & Multiview Attention LSTM  & \begin{tabular}[c]{@{}l@{}}Feature engineering \\ (semantics, affective,\\ incongruity, stylistic \\ and structural)\end{tabular}     & \begin{tabular}[c]{@{}l@{}}F1: 70.4\% (IroSva'19)\\ F1: 95.7\% (Barbieri'15))\\ F1: 96\% (Salas'17)\\ F1: 80.6\% (HAHA'19)\end{tabular} \\ \hline
    
    \citet{Zhang2017InvestigationsRecognition}     & Humour    & 16000 One-Liner (32,002)     & \begin{tabular}[c]{@{}l@{}}Contextual knowledge, \\ Subjectivity, affective polarity, \\ ambiguity, incongruity\end{tabular}     & -    & \begin{tabular}[c]{@{}l@{}}Feature engineering \\ (Contextual knowledge, \\ Subjectivity, polarity, \\ ambiguity, incongruity)\end{tabular}     & F1: 85\%     \\ \hline
    
    \citet{Mann2024CLEFClassification}     & Humour, Pun   & CLEF 2024 JOKER Tasks    & TF-IDF   & \begin{tabular}[c]{@{}l@{}}Logistic regression, \\ Naive Bayes (NB),\\ SVM, MarianMTModel\end{tabular}     &  & \begin{tabular}[c]{@{}l@{}}F1: 83\% (Pun)\\ F1: 61\% (Humour)\end{tabular}    \\ \hline
    
    \citet{DeMarez2024THInC:Detection}       & Humour        & SemEval 2021 Task 7     & \begin{tabular}[c]{@{}l@{}}Polarity, emotions, offense, \\ subjectivity, hate, stance,\\ ambiguity, adult language\end{tabular}     & GA2M   & \begin{tabular}[c]{@{}l@{}}Feature engineering \\ (Polarity, emotions, \\ offense, subjectivity, \\ hate, stance, ambiguity, \\ adult language)\end{tabular} & F1: 85\%      \\ \hline
    
   \citet{Mahajan2024AnModels}        & Humour        & Yelp reviews  & Emotive, incongruity, intensity  & \begin{tabular}[c]{@{}l@{}}Stacking-based ensemble, \\ NB, SVM, MLP,\\ majority-vote ensemble\end{tabular} & \begin{tabular}[c]{@{}l@{}}Feature engineering \\ (Berger’s typology-based\\  incongruity features)\end{tabular}  & F1: 72.57\%  \\ \hline
    
   \citet{Chen2024TalkInterpretation}      & Humour     & Chain-of-humour dataset    & word embeddings   & \begin{tabular}[c]{@{}l@{}}Pre-trained language \\ models (PLMs)\end{tabular}     & Chain-of-thought  & F1: 87\%   \\ \hline
    
  \citet{Abulaish2018Self-DeprecatingApproach}  & Sarcasm   & Twitter API (7 weeks tweet)    & \begin{tabular}[c]{@{}l@{}}Self-referencing, hyperbolic,\\  part-of-speech\end{tabular}  & \begin{tabular}[c]{@{}l@{}}Decision tree, \\ Naive Bayes, \\ Bagging\end{tabular}     &   & F1: 94\%  \\ \hline
    
    \citet{Kamal2020Self-deprecatingApproach}     & Humour        & \begin{tabular}[c]{@{}l@{}}Pun of the Day (4,826), \\ 1600 One-Liner (32.002),\\ Twitter API (20,000)\end{tabular}       & \begin{tabular}[c]{@{}l@{}}Self-deprecating pattern, \\ exaggeration, word embedding,\\ ambiguity, interpersonal eﬀect, \\ phonetic style.\end{tabular}          & Random forest   &    & \begin{tabular}[c]{@{}l@{}}F1: 62\% - 87\% \\ across datasets\end{tabular}  \\ \hline
    
   \citet{KennethOgbuka2024ARecognition}& Humour Styles & Humour styles dataset    & Sentence embedding   & \begin{tabular}[c]{@{}l@{}}NB, Randon forest, \\ XGBoost, DistilBERT\end{tabular}    &    & F1: 78.6\%  \\ \hline
    \end{tabular}
    }
\caption{Summary of Related Works}
\label{Table:related_works}
\end{table}

\end{document}